%% file: main.tex
\pdfoutput=1

\documentclass[11pt]{article}

\usepackage[]{acl}

\usepackage{times}
\usepackage{latexsym}

\usepackage[T1]{fontenc}

\usepackage[utf8]{inputenc}

\usepackage{microtype}

\usepackage{inconsolata}

\usepackage[inkscapelatex=false]{svg}
\usepackage{amsmath}
\usepackage{multirow}
\usepackage{enumitem}

\usepackage{paralist}

\title{Multilingual Contrastive Decoding via Language-Agnostic Layers Skipping}

\author{
    Wenhao Zhu$^{1}$\footnotemark[1]\text{,} \textbf{Sizhe Liu}$^{1}$\footnotemark[1]\text{,} \textbf{Shujian Huang}$^{1}$\text{,} \textbf{Shuaijie She}$^{1}$\text{,} \textbf{Chris Wendler}$^{2}$\textbf{,} \textbf{Jiajun Chen}$^{1}$ \\
    $^{1}$ \text{National Key Laboratory for Novel Software Technology, Nanjing University} $^{2}$ \text{EPFL} \\
    \small\texttt{zhuwh@smail.nju.edu.cn}, \small\texttt{liusz@nju.edu.cn}, \small\texttt{shesj@nju.edu.cn}, \\
    \small\texttt{huangsj@nju.edu.cn}, \small\texttt{chris.wendler@epfl.ch}, \small\texttt{chenjj@nju.edu.cn} \\
}

\begin{document}
\maketitle

\renewcommand{\thefootnote}{\fnsymbol{footnote}}
\footnotetext[1]{Equal contribution.}
\renewcommand{\thefootnote}{\arabic{footnote}}

\input{tex/0-abstract}

\input{tex/1-Introduction}

\input{tex/2-Background}

\input{tex/3-Method}

\input{tex/4-Result}

\input{tex/5-Conclusion}



\bibliography{anthology,custom}

\appendix



\input{tex/6-Appendix}

\end{document}

%% file: tex/0-abstract.tex
\begin{abstract}

Decoding by contrasting layers (DoLa), is designed to improve the generation quality of large language models (LLMs) by contrasting the prediction probabilities between an early exit output (amateur logits) and the final output (expert logits).
However, we find that this approach does not work well on non-English tasks.
Inspired by previous interpretability work on language transition during the model's forward pass, we discover that this issue arises from a language mismatch between early exit output and final output.
In this work, we propose an improved contrastive decoding algorithm that is effective for diverse languages beyond English.
To obtain more helpful amateur logits, we devise two strategies to skip a set of bottom, language-agnostic layers based on our preliminary analysis.
Experimental results on multilingual reasoning benchmarks demonstrate that our proposed method outperforms previous contrastive decoding baselines and substantially improves LLM's chain-of-thought reasoning accuracy across 11 languages\footnote{The project will be available at: \url{https://github.com/NJUNLP/SkipLayerCD}.}.
\end{abstract}

%% file: tex/1-Introduction.tex
\section{Introduction}
Contrastive decoding~\citep{li-etal-2023-contrastive} presents a novel approach to enhance the text generation quality of large language models.
At each inference step, contrastive decoding uses logits generated by an amateur model (usually small) to contrast with the output logits of an expert model (usually large).
This reduces the probability of the expert model to make similar mistakes as the amateur model, thus making the generation content more logical and coherent~\cite{li-etal-2023-contrastive, o2023contrastive, zhao2024enhancing}.
To further eliminate the need of finding an extra amateur LLM, \citet{chuang2023dola} propose DoLa, which uses the expert model's early exit output as amateur logits.

\begin{figure}[t]
\centering
\includegraphics[width=0.49\textwidth]{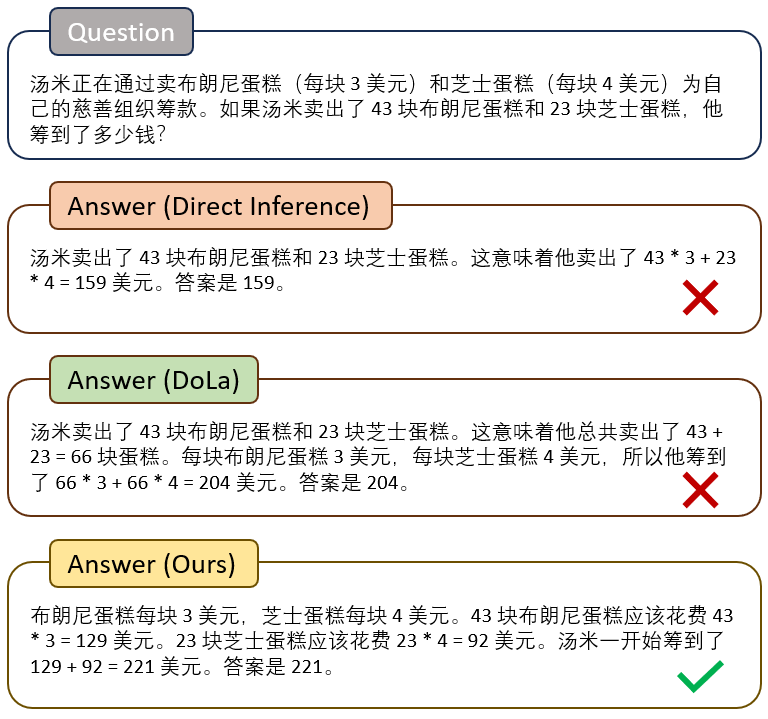}
\caption{Illustration of the superiority of our proposed layer skipping contrastive decoding algorithm over direct inference and DoLa.
}
\end{figure}

However, in this paper, we find that DoLa does not work well on non-English tasks.
Inspired by the recent interpretability study by \citet{wendler2024llamas}, which analyzes the language transitions during the forward pass, we identify that the issue with DoLa arises from the language mismatch between amateur logits and expert logits.
Specifically, the early exit logits accumulate on English tokens even during non-English generation, thus failing to provide a helpful contrastive distribution for the expert model.

\textbf{Contributions} To obtain more helpful amateur logits, we propose an improved contrastive decoding algorithm by skipping a set of lower language-agnostic layers while preserving the computations in the upper transformer blocks (Figures~\ref{fig-method}).
Specifically, we design two strategies to determine the positions for layer skipping: one based on heuristic rules, and the other based on entropy change. 

Our experimental results on multilingual reasoning benchmarks \textsc{mGSM} show that our devised approach significantly outperforms the previous contrastive decoding approach DoLa, and improves the chain-of-thought reasoning accuracy of a group of open-source LLMs: LLaMA2~\cite{llama2}, LLaMA3~\cite{meta2024introducing}, Mistral~\cite{jiang2023mistral}, etc., across 11 languages. 
The performance gap between our approach and DoLa on the multilingual benchmark also validates the findings about the language transition of intermediate decodings across the layers of LLMs by \citet{wendler2024llamas} and provides further insight into the working patterns of LLMs.

%% file: tex/2-Background.tex
\section{Background and Preliminary Analysis}
In this section, we will briefly introduce the background of contrastive decoding and discuss why DoLa can not work well on non-English tasks. 

\subsection{Contrastive Decoding}
While LLMs have shown impressive potential as foundation models~\cite{llama2, jiang2023mistral}, they still easily make logical mistakes or generate hallucinations, especially in scenarios such as complex reasoning.
To improve LLM's generation quality, \citet{li-etal-2023-contrastive} propose an effective decoding algorithm called contrastive decoding.
This method uses the output logits from a small amateur model as a negative bias and subtracts this bias from the output logits of a large expert model at each inference step.
But in practice, it is often hard to find a suitable amateur LLM that is smaller in size and shares the same vocabulary as the expert LLM.
Therefore, \citet{chuang2023dola} propose an amateur-free contrastive decoding method DoLa, which uses the early exit probabilities from the bottom layers as the amateur logits.

\subsection{The Problem with Early Exit During Multilingual Generation}
\label{sec:exp_pre_chinese}
\begin{figure}[htbp]
\centering
\includegraphics[width=0.45\textwidth]{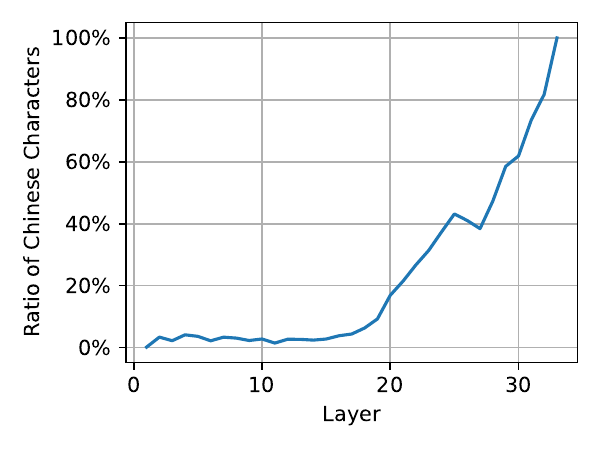}
\caption{The ratio of generating Chinese tokens for each layer of Mistral-7B on solving the \textsc{mGSM} task (Chinese part) with chain-of-thought.}
\label{fig-chinese-ratio}
\end{figure}
Despite DoLa's effectiveness on improving English generation, we discover that this approach does not perform well on non-English tasks~(see Table~\ref{res-mul}).
We find that this issue stems from a language mismatch between the early exit output and the final output.
Figure \ref{fig-chinese-ratio} provides an empirical evidence for this observation\footnote{We explain the detailed setting of Figure~\ref{fig-chinese-ratio} in Appendix~\ref{sec:app_exp_set_pre}.} where we use the logit lens~\cite{LogitLens} to analyze each layer's output of Mistral-7B.
We observe that Mistral does not generate tokens in the target language (Chinese, in this case) in the early exit output until it reaches the last few layers.
This language mismatch fails to contribute to meaningful contrasting, thus leading to DoLa's shortcomings on non-English tasks.

\subsection{LLM's Three-Phase Working Pattern}
\begin{figure*}[ht]
\centering
\includegraphics[width=1.0\textwidth]{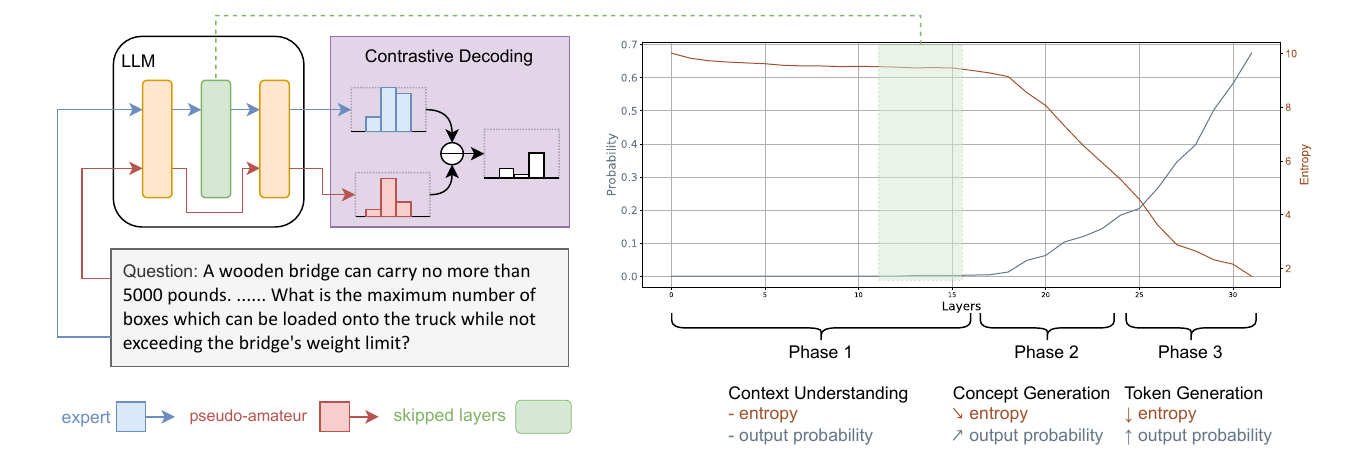}
\caption{Illustration of our devised contrastive decoding approach. The idea of the line chart and three phases division are borrowed from the work of \citet{wendler2024llamas}. In the line chart, ``probability'' denotes the token generation probability and ``entropy'' denotes the entropy of the prediction distribution.}
\label{fig-method}
\end{figure*}

A theoretical explanation for this language mismatch could be attributed to LLM's three-phase working pattern (Figure~\ref{fig-method}).
In the work of \citet{wendler2024llamas}, it is discovered that for simple multilingual in context learning prompts the forward computation of LLMs can be divided into three phases: understanding the context, generating the concept for the next token, and converting the concept into target language tokens. 
Furthermore, during this process, the change in prediction entropy serves as a crucial indicator for each specific working phase.

{
\renewcommand{\arraystretch}{1.1}
\begin{table}[htbp]
    \footnotesize
    \centering
    \begin{tabular}{lcccccc}
\hline
\textbf{Position} & \textsc{DE} & \textsc{FR} & \textsc{ES} & \textsc{RU} & \textsc{ZH} & \textsc{AVG} \\
\hline
{[}4, 8) & 36.0 & 35.2 & 37.6 & 35.2 & 34.0 & 35.6 \\
{[}8, 12) & \textbf{40.0} & 39.2 & \textbf{43.2} & \textbf{38.0} & 34.8 & 39.0 \\
{[}12, 16) & 38.0 & \textbf{39.6} & 40.8 & 37.6 & \textbf{40.0} & \textbf{39.2} \\
{[}16, 20) & 37.2 & 33.6 & 38.4 & 37.2 & 32.0 & 35.7 \\
{[}20, 24) & 34.4 & 33.6 & 38.8 & 35.2 & 34.4 & 35.3 \\
{[}24, 28) & 33.2 & 36.0 & 35.2 & 30.4 & 34.8 & 33.9 \\
\hline
    \end{tabular}
    \caption{\label{pre-mistral-skip-different-layer} We set different positions for layer-skipping during contrastive decoding and observe Mistral-7B's reasoning accuracy on \textsc{mGSM} dataset.} 
\end{table}
}

According to this analysis, the early exit approach skips the final language conversion phase, leading to a language mismatch between amateur logits and expert logits.
Therefore, a promising approach to address this issue is to skip only partial middles layers and complete the computation within the top transformer blocks.
To validate this idea, we conduct a preliminary study (Table~\ref{pre-mistral-skip-different-layer}) by skipping layers at different positions.
The results indicate that skipping the layers in the lower half of the model (context understanding phase) produces more helpful amateur logits, whereas skipping the top layers most significantly degrades performance.

%% file: tex/3-Method.tex
\section{Methodology}
Based on our previous discussion, we now introduce our devised algorithm in detail.
Our intuition is that when the model's computation is perturbed during the context understanding phase, such as by skipping some layers, it tends to generate fluent but unreasonable content due to the poorly extracted features.
This perturbation causes the output distribution to serve as useful amateur logits for contrastive decoding.

{
\renewcommand{\arraystretch}{1.1}
\setlength\tabcolsep{4pt} 

\begin{table*}[htbp]
\footnotesize
\centering
\begin{tabular}{lccccccc|cccc|ccc}
\hline
 & \textsc{EN} & \textsc{DE} & \textsc{FR} & \textsc{ES} & \textsc{RU} & \textsc{ZH} & \textsc{JA} & \textsc{TH} & \textsc{TE} & \textsc{BN} & \textsc{SW} & \textsc{AVG} & \textsc{HRL} & \textsc{LRL} \\
\hline
\multicolumn{15}{c}{\textit{Mistral 7B}} \\
\hline
Direct & 45.6 & 33.2 & 34.4 & 34.8 & 32.0 & 34.4 & 19.2 & 16.4 & 2.0 & 12.0 & 6.0 & 24.5 & 33.4 & 9.1 \\
DoLa & 46.0 & 35.6 & 34.4 & \underline{39.6} & 31.2 & 31.6 & 20.0 & 16.4 & \underline{2.4} & 8.8 & \underline{8.4} & 24.9 (+0.4) & 34.1 (+0.7) & 9.0 (-0.1) \\
SL-H & \underline{50.4} & \underline{38.4} & \underline{36.4} & \textbf{42.0} & \underline{36.4} & \underline{39.6} & \textbf{24.8} & \textbf{19.6} & \textbf{6.8} & \textbf{16.0} & \textbf{10.8} & \textbf{29.2} (+4.7) & \underline{38.3} (+4.9) & \textbf{13.3} (+4.2) \\
SL-D & \textbf{54.0} & \textbf{38.8} & \textbf{38.8} & 39.2 & \textbf{40.4} & \textbf{41.2} & \underline{22.4} & \underline{17.2} & \underline{2.4} & \underline{14.8} & 6.4 & \underline{28.7} (+4.1) & \textbf{39.3} (+5.9) & \underline{10.2} (+1.1) \\
\hline
\multicolumn{15}{c}{\textit{Deepseek 7B}} \\
\hline
Direct & 17.2 & 13.2 & \textbf{16.4} & \textbf{18.0} & 12.0 & 16.8 & 10.0 & 2.4 & 2.0 & \underline{2.8} & 2.0 & 10.3 & 14.8 & 2.3 \\
DoLa & 11.2 & 6.0 & 7.6 & 6.4 & 8.0 & 14.4 & 2.4 & 1.6 & 2.0 & 1.6 & \underline{2.4} & 5.8 (-4.5) & 8.0 (-6.8) & 1.9 (-0.4) \\
SL-H & \underline{20.8} & \underline{17.2} & 12.8 & \underline{14.0} & \underline{14.8} & \textbf{24.0} & \underline{10.4} & \textbf{6.4} & \underline{2.4} & \textbf{3.2} & \textbf{4.0} & \underline{11.8} (+1.6) & \underline{16.3} (+1.5) & \textbf{4.0} (+1.7) \\
SL-D & \textbf{24.0} & \textbf{18.0} & \underline{13.2} & 12.8 & \textbf{15.2} & \underline{22.8} & \textbf{12.4} & \underline{5.6} & \textbf{2.8} & 2.4 & \underline{2.4} & \textbf{12.0} (+1.7) & \textbf{16.9} (+2.1) & \underline{3.3} (+1.0) \\
\hline
\multicolumn{15}{c}{\textit{Baichuan 2 7B}} \\
\hline
Direct & \underline{29.6} & 14.8 & \textbf{23.6} & 20.0 & \underline{16.8} & 27.2 & 10.4 & 8.8 & 1.6 & 4.0 & \underline{3.6} & 14.6 & 20.3 & 4.5 \\
DoLa & 27.2 & \textbf{19.2} & \underline{22.0} & 17.6 & 16.0 & 31.6 & \underline{13.2} & 8.4 & 0.8 & 4.0 & 2.4 & 14.8 (+0.2) & 21.0 (+0.6) & 3.9 (-0.6) \\
SL-H & \underline{29.6} & \underline{18.0} & \underline{22.0} & \underline{23.2} & 15.2 & \underline{32.8} & \textbf{13.6} & \textbf{10.8} & \textbf{2.4} & \underline{4.8} & \textbf{4.8} & \textbf{16.1} (+1.5) & \textbf{22.1} (+1.7) & \textbf{5.7} (+1.2) \\
SL-D & \textbf{30.0} & 14.8 & 18.4 & \textbf{24.8} & \textbf{19.6} & \textbf{34.4} & 11.6 & \underline{9.6} & \textbf{2.4} & \textbf{5.2} & \underline{3.6} & \underline{15.9} (+1.3) & \underline{21.9} (+1.6) & \underline{5.2} (+0.7) \\
\hline
\multicolumn{15}{c}{\textit{LLaMA 3 8B}} \\
\hline
Direct & \underline{57.6} & 41.2 & 40.8 & 48.8 & \underline{47.2} & 42.4 & \underline{31.2} & \textbf{42.0} & 18.0 & 30.0 & 24.0 & 38.5 & 44.2 & 28.5 \\
DoLa & 56.8 & 42.0 & 41.6 & 49.2 & 43.6 & 41.6 & \underline{31.2} & 38.4 & 19.6 & 26.8 & 26.4 & 37.9 (-0.5) & 43.7 (-0.5) & 27.8 (-0.7) \\
SL-H & \textbf{59.2} & \textbf{48.4} & \underline{42.0} & \textbf{55.2} & \textbf{48.4} & \underline{43.6} & \textbf{36.0} & 40.8 & \textbf{26.8} & \textbf{40.8} & \underline{28.0} & \textbf{42.7} (+4.2) & \textbf{47.5} (+3.4) & \textbf{34.1} (+5.6) \\
SL-D & 54.4 & \underline{42.4} & \textbf{43.6} & \underline{50.4} & 46.0 & \textbf{45.6} & \underline{31.2} & \underline{41.6} & \underline{25.6} & \underline{38.4} & \textbf{28.4} & \underline{40.7} (+2.2) & \underline{44.8} (+0.6) & \underline{33.5} (+5.0) \\
\hline
\multicolumn{15}{c}{\textit{LLaMA 2 13B}} \\
\hline
Direct & \underline{34.8} & 21.6 & 22.0 & 26.0 & 20.0 & 20.4 & \underline{13.6} & \underline{6.4} & \underline{1.2} & 1.6 & 2.8 & 15.5 & 22.6 & 3.0 \\
DoLa & 32.8 & 25.2 & \textbf{25.6} & 25.6 & 18.4 & 19.6 & 10.8 & \underline{6.4} & 0.4 & 2.8 & \underline{5.2} & 15.7 (+0.2) & 22.6 (-0.1) & \underline{3.7} (+0.7) \\
SL-H & \textbf{36.4} & \textbf{27.6} & \underline{24.8} & \underline{26.4} & \textbf{22.0} & \underline{23.2} & 12.0 & \textbf{7.6} & \underline{1.2} & \textbf{5.2} & \textbf{6.0} & \underline{17.5} (+2.0) & \underline{24.6} (+2.0) & \textbf{5.0} (+2.0) \\
SL-D & \underline{34.8} & \underline{26.0} & 24.4 & \textbf{30.4} & \underline{21.6} & \textbf{25.2} & \textbf{17.6} & 4.8 & \textbf{1.6} & \underline{4.4} & 3.6 & \textbf{17.7} (+2.2) & \textbf{25.7} (+3.1) & 3.6 (+0.6) \\
\hline

    \end{tabular}
    \caption{\label{res-mul}Comparison results of different decoding methods on \textsc{mGSM}. ``\textsc{HRL}'' and ``\textsc{LRL}'' denote the average performance on seven high-resource languages and four low-resource languages. ``SL-H'' and ``SL-D'' denote the heuristic skipping and dynamic skipping of our approach. The bold and underlined text denotes the best and second best results along the column.}
\end{table*}
}

\subsection{Overview}
To obtain the amateur logits, our approach skips the layer span \([m, n)\) of the model during forward computation.
After performing the layer skipping, the hidden state computation of a transformer model with \(N\) layers can be written as:
\begin{equation}
\begin{aligned}
    h_i &= \left\{
    \begin{aligned}
        \mathtt{Emb}(x) \quad & i = 0 \nonumber \\
        \mathtt{L}_i(h_{i-1}) \quad & i \in (0, m) \cup [n, N] \nonumber \\
        h_m \quad & i \in [m, n) \nonumber 
    \end{aligned}
    \right. \\
    p_a &= f_{out}(h_N) \nonumber
\end{aligned}
\end{equation}
where \(x\) denotes a input token, \(h_i\) represents the output hidden states of the i-th layer and \(L_i\) represents the i-th layer of the model. The indices
\(m\) and \(n\) mark the beginning and end of the layer skipping span.
In the last layer, the prediction probability of amateur \(p_a\) can be obtained by grounding the hidden state $h_N$ into the output embedding space via the output function \(f_{out}\).

\subsection{Strategies for Layer Skipping}
Correctly setting the skipping span \([m, n)\) is non-trivial.
We propose the following two strategies for position selection: 

\noindent\paragraph{Heuristic layer skipping (SL-H)} 
Considering that the first phase involves the bottom half of the layers in LLMs, a basic strategy is to randomly skip a few layers in this region, excluding the very bottom layers that process low-level lexical information. 
For each evaluated sample, 
we sample the begin-skipping layer $m$ and set the end-skipping layer $n$ with following equations:
\begin{equation}
\begin{aligned}
m &= m \sim \mathcal{U}(4, N/2-1) \\
n &= m+\text{round}(N/8) \\
\end{aligned}
\end{equation}

\noindent\paragraph{Dynamic layer skipping (SL-D)} 
To automatically determine the skipping position, we propose a dynamic algorithm based on entropy change.
This approach is motivated by the sharp decrease in entropy that occurs between the first and second phase (Figure~\ref{fig-method}). 
Specifically, we calculate the entropy of the output distribution for each layer and identify the position where the entropy decreases by more than a predefined threshold \(\delta\). We set the end of the skipping span $n$ to this position (Figure \ref{fig-method}). 
The formal description of this algorithm is as follows:
\begin{equation}
\begin{aligned}
m &= n - \text{round}(N/8) \\
n &= \min \limits_{i \in \{k, \ldots, N\}} \{ i \colon e'_i - e'_{i-1} > \delta \}
\end{aligned}
\end{equation}
where \(e_i\) is the entropy of output probability of i-th layer computed by \(-\sum f_{out}(h_i) \log f_{out}(h_i)\), and \(k\) is the minimum index that ensures \(m > 6\), for excluding early layers\footnote{
In practical applications, to stabilize spikes or fluctuations in the entropy value \(e_i\), we implement average pooling by computing \(e'_i=(e_i + e_{i+1})/2\). Additionally, we enforce a descending constraint such that \(\forall j > n, e'_j < e'_n\).
}.

%% file: tex/4-Result.tex
\vspace{-0.2cm}
\begingroup
\setlength\tabcolsep{5pt} 
\renewcommand{\arraystretch}{1.1} 
\begin{table}[htbp]
    \centering
    \footnotesize
    \begin{tabular}{lcccc}
\hline
 & \textsc{AQuA}  & \textsc{GSM8K} & \textsc{GSM-Plus}  & \textsc{AVG} \\
\hline
\multicolumn{5}{c}{\textit{Mistral 7B}} \\
\hline
Direct & 32.3 & 43.6 & 34.7 & 36.9 \\
DoLa & 29.1 & 46.6 & 35.0 & 36.9 (+0.0) \\
SL-H & \underline{35.8} & \underline{49.0} & \underline{36.9} & \underline{40.6} (+3.7) \\
SL-D & \textbf{37.0} & \textbf{50.6} & \textbf{37.4} & \textbf{41.6} (+4.8) \\
\hline
\multicolumn{5}{c}{\textit{Deepseek 7B}} \\
\hline
Direct & 24.4 & 15.0 & 11.9 & 17.1 \\
DoLa & 14.6 & 10.2 & 7.1 & 10.6 (-6.5) \\
SL-H & \underline{26.4} & \underline{18.9} & \textbf{13.7} & \underline{19.6} (+2.5) \\
SL-D & \textbf{28.7} & \textbf{20.3} & \underline{13.2} & \textbf{20.8} (+3.6) \\
\hline
\multicolumn{5}{c}{\textit{Baichuan 2 7B}} \\
\hline
Direct & 25.2 & 20.3 & 16.0 & 20.5 \\
DoLa & \underline{28.0} & 22.0 & 15.4 & 21.8 (+1.3) \\
SL-H & \underline{28.0} & \underline{23.3} & \underline{17.8} & \underline{23.0} (+2.5) \\
SL-D & \textbf{29.5} & \textbf{23.6} & \textbf{18.0} & \textbf{23.7} (+3.2) \\
Vanilla & 26.0 & 20.9 & 15.8 & 20.9 (+0.4) \\
\hline
\multicolumn{5}{c}{\textit{LLaMA 3 8B}} \\
\hline
Direct & 34.6 & \underline{55.4} & \textbf{43.9} & 44.6 \\
DoLa & 37.4 & 49.6 & 37.0 & 41.3 (-3.3) \\
SL-H & \textbf{43.3} & 53.3 & \underline{41.1} & \textbf{45.9} (+1.3) \\
SL-D & \underline{38.2} & \textbf{56.0} & 40.7 & \underline{44.9} (+0.3) \\
\hline
\multicolumn{5}{c}{\textit{LLaMA 2 13B}} \\
\hline
Direct & 25.6 & 26.0 & 20.8 & 24.1 \\
DoLa & 29.1 & 27.7 & 19.5 & 25.5 (+1.3) \\
SL-H & 28.3 & 28.0 & 21.7 & 26.0 (+1.9) \\
SL-D & \textbf{31.5} & \underline{30.5} & \underline{22.1} & \underline{28.0} (+3.9) \\
Vanilla & \underline{29.9} & \textbf{32.1} & \textbf{23.4} & \textbf{28.5} (+4.4) \\
\hline
    \end{tabular}
    \caption{\label{res-eng}Comparison results of different decoding method on English reasoning benchmarks.}
\end{table}
\endgroup

\section{Experiments}
\subsection{Settings}

\noindent\paragraph{Expert LLMs}
In our experiments, we consider several popular LLMs with with different model sizes, including Mistral-7B~\citep{jiang2023mistral}, Baichuan2-7B~\citep{baichuan2023baichuan2}, Deepseek-7B~\citep{deepseek-llm}, LLaMA3-8B~\citep{meta2024introducing} and LLaMA2-13B~\citep{touvron2023llama}. 
During decoding, we use few-shot chain-of-thought prompting.
More details are reported in Appendix \ref{sec:app_exp_set}.

\noindent\paragraph{Baseline decoding algorithms}
We include three key baselines for comparison: 
\begin{compactenum}
\item \textbf{Direct}: direct inference without using contrastive decoding. 
\item \textbf{DoLa}~\citep{chuang2023dola}: the latest amateur-free contrastive decoding variant that uses early exit instead of completing the computation within the top transformer blocks.
\item \textbf{Vanilla}~\cite{li-etal-2023-contrastive}: vanilla contrastive decoding approach that requires both expert model and an extra amateur model\footnote{Note that only for Baichuan2-7B and LLaMA2-13B, we can find suitable amateur models: Baichuan's 220B token checkpoint and Sheared-LLaMA-1.3B from \citet{xia2023sheared} respectively, to implement vanilla contrastive decoding.}. 
\end{compactenum}

\noindent\paragraph{Evaluation datasets}
We consider both multilingual reasoning tasks \textsc{mGSM}\footnote{The \textsc{mGSM} dataset contains English, six other high-resource languages and four low-resource languages.}~\citep{shi2022language} 
and
English reasoning benchmarks \textsc{AQuA}~\citep{ling-etal-2017-program}, \textsc{GSM8K}~\citep{cobbe2021gsm8k} and \textsc{GSM-Plus}~\citep{li2024gsm}\footnote{We evaluate on the subset of GSM-Plus where the answer is an integer.} to evaluate the effectiveness of our proposed method.

\subsection{Results}
\noindent\paragraph{Layer skipping provides helpful distribution for contrastive decoding}
In both Table \ref{res-mul} and Table \ref{res-eng}, we can see that our approach (\textsc{SL-H} \& \textsc{SL-D}) outperforms direct inference by a large margin in average.
The dynamic layer skipping approach achieves better performance than heuristic layer skipping in most of benchmark results.
These results demonstrate that our layer skipping approach provides helpful amateur logits for contrastive decoding.

\noindent\paragraph{Our devised approach enjoys more superiority in multilingual scenarios}
As shown in Table \ref{res-mul}, our proposed approach improves the reasoning accuracy of all evaluated LLMs over baseline decoding algorithms.
These results, especially the failure of DoLa in multilingual tasks demonstrates the importance of keeping top transformer layers during contrastive decoding.

To further illustrate the shortcomings in the design of DoLa, we present an ablation study in which we skip the computation both in the selected region [m, n) and in the remaining layers [n, N]. 
Experimental results in Appendix~\ref{sec:ablation} show that maintaining the computation in the top layers is essential. 

\noindent\paragraph{Our proposed approach eliminates the need for finding an extra amateur model for contrastive decoding}
Compared to the vanilla contrastive decoding approach, our proposed approach does not require an extra amateur model while achieving comparable performance (Table~\ref{res-eng}).
This makes it more applicable in practical scenarios.

%% file: tex/5-Conclusion.tex
\section{Conclusion}
This paper is motivated by the observation of the failure of contrastive decoding variant DoLa on multilingual generation.
Through empirical analysis and drawing inspiration from previous interpretability study on LLM's three-phase working pattern, we find that the failure stems from a language mismatch between the early exit output and the final output.
To address this issue, we propose an improved contrastive decoding algorithm by skipping a set of lower layers and preserving the computation in the upper transformer blocks, which are essential for language transition.
Experimental results on both multilingual and monolingual benchmarks demonstrate the effectiveness of our method.

\section*{Limitations}
Below we discuss potential limitations of our work:
\begin{compactenum}
\item Extra inference cost: although contrastive decoding algorithms improve the generation quality of LLMs, they introduces additional computational cost for obtaining amateur logits. This results in a slower inference speed.
\item Limited model range: we conducted experiments using several popular LLMs, but the range of considered models may still be limited. For instance, we have not considered LLMs with the Mixture-of-Experts architecture.
\end{compactenum}

\section*{Acknowledgement}
Shujian Huang is the corresponding author. 
This work is supported by National Science Foundation of China (No. 62376116, 62176120).
Wenhao Zhu is also supported by China Scholarship Council (No.202306190172).

%% file: tex/6-Appendix.tex
\clearpage
\section{Details of Preliminary Analysis}
\label{sec:app_exp_set_pre}
For the preliminary analysis in Section~\ref{sec:exp_pre_chinese}, we evaluate the Mistral 7B model on the MGSM dataset in Chinese.
During generation, we apply the early exit approach to each layer to obtain the output probabilities for the vocabulary and observe the top-1 token from each layer. 
Subsequently, we examine whether each token from each layer is a Chinese character and compute the ratio of such tokens over all generation positions where the output token of the final layer is a Chinese character.

\section{Detailed Experiment Setting}
\label{sec:app_exp_set}
We use the multilingual examples from \citet{shi2022language} to evaluate on the multilingual MGSM dataset. 
Specifically, we use 2-shot examples for Telugu, 4-shot examples for Bengali and Thai, and 8-shot examples for the remaining languages.
For other datasets, we use the English examples from \citet{NEURIPS2022_9d560961}, 4-shot for \textsc{AQuA}, 8-shot for \textsc{GSM8K} and \textsc{GSM-Plus}.

We use the modified contrastive decoding from \citet{o2023contrastive} for both of our approaches and the vanilla contrastive decoding with \(\alpha=0.1, \beta=0.5\).
We set \(\delta=0.1\) for the dynamic layer skipping.
We use greedy decoding for all evaluated methods.
Regarding the implementation of DoLa, we follow the implementation of the original paper~\citep{chuang2023dola} and select the early exit layer in the lower half layers.

\section{Dataset Statistics}
We use the development set of AQuA and test set of other datasets for evaluation. The dataset statistics are reported in Table \ref{data-size}.

{
\renewcommand{\arraystretch}{1.1}
\begin{table}[ht]
    \footnotesize
    \centering
    \begin{tabular}{ccc}
\hline
\textbf{Dataset} & \textbf{\# Lang} & \textbf{\# Sample} \\
\hline
\textsc{AQuA} & 1 & 254 \\
\textsc{GSM8K} & 1 & 1,319 \\
\textsc{GSM-Plus} (subset) & 1 & 8,651 \\
\textsc{mGSM} & 11 & 2,750 \\
\hline
    \end{tabular}
    \caption{\label{data-size}Dataset statistics of our used datasets.}
\end{table}
}

\section{Ablation Study}
\label{sec:ablation}
To further illustrate the shortcomings in the design of DoLa, we present an ablation study in which we skip the computation both in the selected region [m, n) and in the remaining layers [n, N]. 
The modified versions are called SL-H (E) and SL-D (E) in Table~\ref{res-mul-sle} .
Experiment results show that SL-H/D (E) performs worse than our proposed method (SL-H/D), demonstrating the necessity of maintaining computations in the top layers.

{
\renewcommand{\arraystretch}{1.1}
\setlength\tabcolsep{4pt} 
\begin{table*}[ht]
\footnotesize
\centering
\begin{tabular}{lccccccc|cccc|c}
\hline
 & \textsc{EN} & \textsc{DE} & \textsc{FR} & \textsc{ES} & \textsc{RU} & \textsc{ZH} & \textsc{JA} & \textsc{TH} & \textsc{TE} & \textsc{BN} & \textsc{SW} & \textsc{AVG} \\
\hline
Direct & 45.6 & 33.2 & 34.4 & 34.8 & 32.0 & 34.4 & 19.2 & 16.4 & 2.0 & 12.0 & 6.0 & 24.5 \\
DoLa & 46.0 & 35.6 & 34.4 & 39.6 & 31.2 & 31.6 & 20.0 & 16.4 & 2.4 & 8.8 & 8.4 & 24.9 (+0.4) \\
SL-H & 50.4 & 38.4 & 36.4 & 42.0 & 36.4 & 39.6 & 24.8 & 19.6 & 6.8 & 16.0 & 10.8 & 29.2 (+4.7) \\
SL-H (E) & 46.4 & 33.2 & 37.6 & 34.0 & 30.8 & 33.2 & 17.2 & 14.0 & 1.6 & 12.0 & 7.2 & 26.7 (+2.2) \\
SL-D & 54.0 & 38.8 & 38.8 & 39.2 & 40.4 & 41.2 & 22.4 & 17.2 & 2.4 & 14.8 & 6.4 & 28.7 (+4.1) \\
SL-D (E) & 45.2 & 30.8 & 34.0 & 34.4 & 27.6 & 30.4 & 18.8 & 14.0 & 0.8 & 12.4 & 4.0 & 22.9 (-2.6) \\
\hline

    \end{tabular}
    \caption{\label{res-mul-sle}Accuracy of early exit variant on MGSM with Mistral 7B.}
\end{table*}
}

\section{Used Scientific Artifacts}
Below lists scientific artifacts that are used in our
work. For the sake of ethic, our use of these artifacts is consistent with their intended use.
\begin{compactenum}
    \item \textit{Transformers (Apache-2.0 license)}, a framework that provides thousands of pretrained models to perform tasks on different modalities such as text, vision, and audio.
\end{compactenum}